\documentclass{article}
\usepackage{blindtext}

\usepackage{natbib}

\usepackage[utf8]{inputenc}
\usepackage[english]{babel}
\usepackage{authblk}

\usepackage{lastpage}

\usepackage{url}

\usepackage{graphicx}
\usepackage{booktabs}
\usepackage[a4paper, total={6.5in, 9in}]{geometry}
\usepackage{blindtext}
\usepackage{longtable}
\usepackage{pdflscape}
\usepackage{comment}
\usepackage{multirow}
\usepackage{array}
\usepackage{multicol}%
\usepackage{comment}%
\usepackage{subcaption}%
\usepackage{caption}
\usepackage{xurl}
\usepackage{hyperref}

\usepackage{caption}
\usepackage{algorithm}
\usepackage{algpseudocode}

\usepackage{xcolor}
\hypersetup{
    colorlinks=true,
    linkcolor=blue,
    filecolor=blue,
    urlcolor=blue,
    citecolor=black,
    }

\title{Uncovering Fairness through Data Complexity as an Early Indicator}

\author[1]{Juliett Suárez Ferreira}
\author[2]{Marija Slavkovik}
\author[1,3]{Jorge Casillas}
\affil[1]{Data Science and Computational Intelligence Institute (DaSCI), University of Granada.}
\affil[2]{Department of Information Science and Media Studies, University of Bergen.}
\affil[3]{Department of Computer Science and Artificial Intelligence (DECSAI), University of Granada.}
\date{}

\begin{document}
\maketitle

\begin{abstract}
Fairness constitutes a concern within machine learning (ML) applications. Currently, there is no study on how disparities in classification data complexity between privileged and unprivileged groups could influence the fairness of solutions, which serves as a preliminary indicator of potential unfairness. In this work, we investigate this gap, specifically, we focus on synthetic datasets designed to capture a variety of biases ranging from historical bias to measurement and representational bias to evaluate how various complexity metrics differences correlate with group fairness metrics. We then apply association rule mining to identify patterns that link disproportionate complexity differences between groups with fairness-related outcomes, offering data-centric indicators to guide bias mitigation. Our findings are also validated by their application in real-world problems, providing evidence that quantifying group-wise classification complexity can uncover early indicators of potential fairness challenges. This investigation helps practitioners to proactively address bias in classification tasks.

\end{abstract}

\section{Introduction}

Artificial intelligence (AI), particularly machine learning (ML) classification models, have become integral to modern decision-making processes, significantly impacting individuals and communities. While these systems offer unprecedented capabilities, they also present critical challenges regarding fairness and non-discrimination. ML-based systems often perpetuate or amplify existing biases \citep{EUPunderstanding_2019}.

The scientific community has increasingly focused on fairness metrics and mitigation strategies to measure and minimize the effect of bias and discrimination on ML-based systems; however, there remains a need for a deeper understanding of how classification complexity intertwines with unfairness manifestations. Classification complexity refers to properties of a dataset that make classification harder, including class overlap, imbalanced distributions, high non-linearity, or excessive sparsity. The complexity of classification problems presents a unique challenge in attaining optimal outcomes within AI systems. Understanding the complexity of solving a problem can help practitioners implement targeted pre-processing strategies, select appropriate model architectures, and establish realistic performance expectations. 

While research has advanced in developing debiasing interventions for the pre-processing phase of ML pipelines\footnote{A ML pipeline encompasses data collection, pre-processing, model training, evaluation, and deployment stages}, the relationship between problem complexity and fairness remains under explored. Understanding this relationship in the early stages of the development of ML models could be helpful, as discrimination can manifest itself from this stage, influenced by data collection methods, sampling procedures, and measurement processes \citep{Frederik2018}.

In this study, we investigate the relationship between variations in classification complexity between demographic groups (privileged and unprivileged) and fairness outcomes, operating under the hypothesis that differences in classification complexity among these groups lead to unfairness. We specifically focus on synthetic data scenarios designed to capture a variety of biases ranging from historical bias to measurement and representational bias to evaluate how different complexity metrics correlate with group fairness metrics. We also expand our analysis to real-world datasets, exploring whether the relationships we discover in synthetic experiments carry over to practical cases.

Our main research questions are as follows:

\begin{enumerate}
    \item How do distinct bias-generating mechanisms affect the complexity of classification tasks for privileged versus unprivileged groups?
    \item What measurable connections exist between subgroup complexity differences and fairness metrics?
\end{enumerate}

To address these questions, we generate and analyze 73 synthetic datasets, each representing distinct classification scenarios with varying bias trade-offs. We compute complexity metrics differences between privileged and unprivileged groups and fairness metrics under multiple classification algorithms. We then apply association rule mining to identify patterns that link disproportionate complexity differences with fairness outcomes. For external validation, we apply the same rules in 30 real-world datasets from diverse application domains.

The contributions presented in this paper include:

\begin{itemize}
    \item A systematic exploration of bias parameters (historical, representational, and measurement) to illustrate how each influences complexity differences and subsequent model fairness.
    \item A novel application of complexity difference analysis between privileged and unprivileged groups yields clearer insights on how group-specific intricacies can lead to unfairness.
    \item Empirical results that demonstrate consistent relationships between some complexity metrics differences and fairness metrics, offering data-centric indicators to guide bias mitigation.
\end{itemize}

Although bias patterns may not always be immediately apparent in real-world datasets, our findings provide practitioners with valuable insights for anticipating and addressing fairness challenges early in the pre-processing stage of the ML pipeline.

The remainder of this paper is organized as follows. Section \ref{lab:related} reviews the relevant literature and introduces the fairness and complexity metrics that underlie our analysis. Section \ref{lab:methodology} details our experimental approach, followed by the results in Section \ref{lab:results} and their application to real-world problems in Section \ref{lab:real_world} . We discuss implications in Section \ref{lab:discussion} and conclude in Section \ref{lab:conclusions}.

\section{Background and related work}
\label{lab:related}

This section studies the existing body of literature on fairness metrics, classification complexity, and bias intervention strategies that forms the foundation of our research. We have made particular emphasis on pre-processing methods specifically for binary classification tasks. 

In classification problems, we predict a target variable $Y$ (typically discrete or binary) based on a set of attributes $X$. During classification, the true value of $Y$ is unknown and is instead predicted by observing the attributes $X$ and generating an estimated label $\widehat{Y}$ using a classifier $f$ ($\widehat{Y}=f(X)$). Specifically, our focus will be on binary classification problems where $Y$ has two possible labels: one is considered favorable (positive) and the other unfavorable (negative). For simplicity, we assume that one of the attributes in $X$ is a binary protected attribute that divides the population into two groups: a privileged group with systematic historical advantages and an unprivileged group that has faced disadvantages.

Unfairness occurs when the classifier tends to assign more favorable labels or to incur more errors towards one group, indicating that the classifier's decisions are influenced by the protected attribute. The field of fair AI encompasses both the quantification and mitigation of unfairness across the machine learning pipeline \citep{Feuerriegel2020379, Makhlouf2021, Mehrabi2021}. Two primary concepts of fairness have emerged: individual fairness, which requires similar predictions for similar individuals \citep{Dwork2012}, and group fairness, which demands equivalent predictions across different demographic groups \citep{barocas-hardt-narayanan}. Our work focuses on group fairness within classification tasks.

Pre-processing interventions address fairness by transforming the feature space to achieve independence from sensitive attributes \citep{barocas-hardt-narayanan}. \cite{Canton2024} make a detailed study of these approaches that target discriminatory patterns within the distribution of protected attributes of the data. However, there are additional interventions in the pre-processing phase that help practitioners address the classification challenge, one of which involves considering classification complexity metrics to understand the inherent difficulty of the classification task and identifying potential pre-processing strategies needed or identifying the best technique to solve it.


Recent work has explored the relationship between fairness and class balance \citep{Dablain2024, deng2022}. Moreover, \cite{Brzezinski2024} examines the impact of the proportions of class and protected groups on fairness metrics, enumerating some properties of fairness metrics within these contexts. Although imbalance metrics such as C1 and C2 \citep{Lorena2019} quantify class disparities in classification tasks, research suggests that imbalance primarily amplifies other data characteristics rather than acting as a standalone problem \citep{Batista2004}.

To our knowledge, there has been no work studying how other complexity metrics for classification problems can influence the fairness of the solutions. To fill this gap, our study builds on the work on classification complexity metrics \citep{Lorena2019} and the group fairness metrics for classification problems \citep{Calders2009, Hardt2016, Chouldechova2017}, with the goal of finding relationships between the differences in complexity between privileged and unprivileged groups and the fairness of the solution.

The following subsections dive into key aspects relevant to our study. The subsection \ref{fairness_metrics} will explore various metrics used to evaluate fairness between different groups within a dataset, particularly in the context of binary classification tasks. The subsection \ref{complexity_metrics} will examine different metrics used to determine the complexity of classification tasks, focusing on how factors such as class imbalance, feature overlap, and other dataset characteristics can influence the difficulty of classification. Together, these subsections provide the foundational concepts necessary to understand our study's intersection of fairness and data complexity.

\subsection{Group fairness metrics for classification problems}
\label{fairness_metrics}

Numerous mathematical definitions and analyses of fairness metrics have been presented in the literature \citep{pessach2022,Mehrabi2021,Verma20181}. However, there is no common consensus on which fairness metrics are essential to ensure algorithmic fairness; In addition, incompatibilities have been demonstrated between them \citep{Corbett-Davies}. 

We will focus on group fairness metrics, specifically proposed for classification problems. \cite{barocas-hardt-narayanan} have made an extensive study of fairness metrics for classification. They analyzed the metrics and proposed their closest relative among three non-discrimination criteria: independence, separation, and sufficiency.

Most group fairness metrics can be defined using the entries of the confusion matrix for binary classification problems characterizing the relations between the actual labels ($Y$) and the predicted labels $\widehat{Y}$ by the values True positive (TP) and True negative (TN) denoting the number of instances classified correctly for the positive and the negative class while the False positive (FP) False negative (FN) are the number of misclassified instances for the negative and positive class, respectively. The values are defined for the privileged and unprivileged groups as is reflected in Table \ref{tab:confusionmatrix}.

\begin{table}[ht]
    \caption{Confusion matrix for unprivileged ($U$) and privileged ($P$) groups. $_{tp}$, $_{fp}$, $_{tn}$ and $_{fn}$ are the True Positives, False Positives, True Negatives and False Negatives values for each group.}
    \label{tab:confusionmatrix}
    \centering
    \def\arraystretch{1.5}
    \begin{tabular}{ l  l | c | c | c } 
    & & \multicolumn{2}{c}{Predicted} & \\
    & Actual & Positives & Negatives & Total \\ 
    \hline
    \multirow {3}{*}{Unprivileged} & Positives & $U_{tp}$ & $U_{fn}$ & \\
                                   & Negatives & $U_{fp}$ & $U_{tn}$ & \\
                                   & Total & $U_{p}$ & $U_{n}$ & $U$\\                               
    &&&\\
    \multirow {3}{*}{Privileged} & Positives & $P_{tp}$ &  $P_{fn} $ & \\
                                 & Negatives & $P_{fp}$ &  $P_{tn} $ & \\
                                 & Total & $P_{p}$ & $P_{n}$ & $P$ \\
    \end{tabular}
    
\end{table}

We are using a mathematical definition of each criterion:

\begin{itemize}

    \item  \textbf{Statistical Parity} (\textbf{SP}): Also known as Demographic Parity, stands that the likelihood of a positive outcome (favorable label) should be the same regardless of whether the person is in the protected group. This metric represents the independence criteria and was proposed by \cite{Calders2009}.
    \begin{equation}
    \label{eq_SP}
        \frac{U_{tp} + U_{fp}}{U} = \frac{P_{tp} + P_{fp}}{P}
    \end{equation}

        \item \textbf{Equal Opportunity} (\textbf{EO}): defined as the equality of the true positive rate ($\frac{TP}{TP+FN}$) between the unprivileged and privileged groups. This metric, proposed by \cite{Hardt2016}, is a relaxation of the separation criteria.
    \begin{equation}
        \label{eq_EOpp_TPR}
         \frac{U_{tp}}{U_{tp}+U_{fn}} = \frac{P_{tp}}{P_{tp}+P_{fn}} 
    \end{equation}

    \item \textbf{Predictive Parity} (\textbf{PP}): as is defined in \cite{Chouldechova2017}, privileged and unprivileged groups must have the same false discovery rate ($\frac{FP}{TP+FP}$). This metric satisfies the sufficiency criterion. The positive predictive value is usually referred to as precision and represents the probability of a correct positive prediction.
    \begin{equation}
    \label{eq_PP_FDR}
        \frac{U_{fp}}{U_{tp}+U_{fp}} = \frac{P_{fp}}{P_{tp}+P_{fp}}
    \end{equation}

\end{itemize}

These definitions are typically presented as the difference between their results for the unprivileged and privileged groups, leading to a value that shows unfair treatment of the unprivileged group if it is negative. The range of the three metrics we use is $[-1,1]$. We follow the implementations described by \cite{Bellamy}, which also regard the interval $[-0.1,0.1]$ as fair. 

\subsection{Data complexity metrics for classification problems}
\label{complexity_metrics}


Data complexity metrics for classification problems provide insight into the difficulty of separating data points into their respective classes, guiding the selection and development of appropriate algorithms and pre-processing techniques. Several classification complexity metrics have been proposed, the survey made by \cite{Lorena2019} categorizes the metrics according to different criteria:

\begin{itemize}
    \item Feature-based Metrics: Evaluate the discriminative power of individual features.
    \item Linearity Metrics: Assess whether classes can be separated by linear boundaries.
    \item Neighborhood Metrics: Capture the local structure and class overlap.
    \item Network Metrics: Model the dataset as a graph to extract structural information.
    \item Dimensionality Metrics: Indicate data sparsity based on feature dimensions.
    \item Class Imbalance Metrics: Consider the ratio of examples between different classes.
\end{itemize}

The measurement of complexity has several key implications in the field of machine learning. It plays a crucial role in algorithm selection, as complexity metrics help determine the most appropriate machine learning algorithms; for example, high non-linearity datasets may be better suited for non-linear models such as neural networks or ensemble methods; furthermore, performance prediction benefits from complexity metrics by allowing more accurate predictions of how different classifiers may perform, thus setting pragmatic forecasts and highlighting potential challenges in the learning process \citep{CANO2013}. In terms of data pre-processing, recognizing high complexity can indicate, for example, the necessity for data transformation techniques such as feature extraction \citep{Okimoto2019, SARBAZIAZAD2021}, or addressing class imbalances \citep{BARELLA202183} prior to the application of any classifier. 

Using the differences in complexity metrics between privileged and unprivileged groups, we quantify how much more complex it is to solve the same problem for different groups. This quantification gives us an idea of problems that the classifier may have to correctly assign the classes to different groups if one group is more complex to predict than the other.

Table \ref{tab:complexity_measures} shows a summary of the classic data complexity metrics designed for classification problems, including their category according to \cite{Lorena2019}, the name, the short name we will use in the text (SN), and the interval of each measure (R).

\begin{table}[ht]
    \centering
    \caption{Summary of data complexity metrics for classification problems. The columns contain the category, the name, the short name (SN) and the range (R) of each measure.}
    \label{tab:complexity_measures}
    \begin{tabular}{llcc}
        \toprule
        \textbf{Category} & \textbf{Name} & \textbf{SN} & \textbf{R}\\ 
        \midrule
        \multirow{5}{*}{Feature-based}  & Maximum Fisher’s discriminant ratio & F1 & $(0, 1]$  \\ 
        & Directional vector maximum Fisher’s discriminant ratio & F1v & $(0, 1]$ \\ 
        & Volume of overlapping region & F2 & $[0,1]$  \\ 
        & Maximum individual feature efficiency & F3 & $[0,1]$   \\ 
        & Collective feature efficiency & F4 & $[0,1]$  \\ 
        \midrule
        \multirow{3}{*}{Linearity} & Sum of the error distance by linear programming & L1 & $[0, 1)$  \\ 
        & Error rate of linear classifier & L2 & $[0,1]$  \\ 
        & Non-linearity of a linear classifier & L3 & $[0,1]$ \\ 
        \midrule
        \multirow{6}{*}{Neighborhood} & Fraction of borderline points & N1 & $[0,1]$  \\ 
        & Ratio of intra/extra class nearest neighbor distance & N2 & $[0, 1)$  \\ 
        & Error rate of the nearest neighbor classifier & N3 & $[0,1]$ \\ 
        & Non-linearity of the nearest neighbor classifier & N4 & $[0,1]$\\ 
        & Fraction of Hyper spheres covering the data & T1 & $[0,1]$ \\
        & Local Set Average Cardinality the data & LSC & $[0,1\frac{1}{n}]$ \\
        \midrule
        \multirow{3}{*}{Network} & Average density of the network & density & $[0, 1]$  \\ 
        & Clustering coefficient & cls\_coef & $[0,1]$ \\ 
        & Hub score & hubs & $[0,1]$ \\ 
        \midrule
        \multirow{3}{*}{Dimensionality} & Average number of features per dimension & T2 & $(0, 1]$  \\ 
        & Average number of PCA dimensions per points & T3 & $(0, 1]$ \\ 
        & Ratio of the PCA dimension to the original dimension & T4 & $[0,1]$ \\ 
        \midrule
        \multirow{2}{*}{Class Imbalance} & Entropy of class proportions & C1 & $[0,1]$  \\ 
        & Imbalance ratio & C2 & $[0,1]$  \\ 
        \bottomrule
    \end{tabular}
\end{table}

\section{Methodology}
\label{lab:methodology}

The methodology we follow to complete the analysis of our research questions comprises different parts that we define in this section. First, we generate 73 synthetic datasets of binary classification problems representing various bias scenarios using the framework proposed by \cite{Baumann2023}; the specific details will be detailed in Section \ref{subsec:datasets_generation}. Then, we characterize these datasets using the differences in complexity metrics of the classification task between privileged and unprivileged groups, and the fairness metrics for the solution of each task, addressing each problem using three distinct machine learning methodologies (Logistic Regression, Decision Trees, and K-Nearest Neighbors), each selected to represent diverse learning paradigms. A description of these characterizations is provided in Section \ref{subsec:dataset_characterization}. 

After characterizing all datasets we apply association rules to identify relationships between complexity differences and fairness; the description of the approach followed is explained in Section \ref{subsec:association_rules}. This methodology allows us to analyze how different types of bias influence differences in complexity between privileged and unprivileged groups and, subsequently, how these complexity differences affects achievable fairness outcomes.

\subsection{Datasets generation}
\label{subsec:datasets_generation}

\cite{Baumann2023} proposed a framework to simulate various bias-generating mechanisms. The framework distinguishes between biases that affect the data (user to data) and those that directly affect the solution (data to algorithm). The authors modeled different types of bias to achieve a representation that connects sensitive attributes ($A$), resources ($R$, which denote attributes characterizing individual variables directly relevant to the target variable), and other relevant attributes ($Q$, which represent additional variables that could be relevant to predict the target $Y$ and / or could be influenced by $R$ or the sensitive attribute $A$), leading to the final decision $Y$. The modeled biases are enumerated as follows:

\begin{itemize}
    \item Historical Bias: Occurs when data reflect existing social or structural inequalities, leading to biased outcomes based on sensitive characteristics such as gender or race.
    \item Measurement Bias: Emerges when the proxies used to measure variables are flawed, leading to systematic favoritism or discrimination against certain groups.
    \item Representation Bias: Arises when certain subgroups are underrepresented in the data, causing biased outcomes in decision-making processes.
    \item Omitted Variable Bias: The results of excluding relevant variables in the data set, which can lead to spurious correlations and biased results.
\end{itemize}

Table \ref{tab:syn_ds} presents the description of the synthetic datasets generated with different bias scenarios. The total number of generated datasets is 73, with S1A being the one without any associated bias. Scenario S1C only generates one variation of the S1A scenario with the value of R omitted. The rest of the datasets were generated with different values of the corresponding parameter and are labeled with the respective index of the array in the column Values of Table \ref{tab:syn_ds} (Example: the combinations of S1F are S1F1, S1F2, S1F3 and S1F4 for the values 0.3, 0.5, 0.7 and 0.9 respectively).

\begin{table}[ht]
\centering
\caption{Description of the the synthetic datasets generated with different bias scenarios. Each row contain and scenario described by the abels of the datasets (Label), their description (Explanation), the parameter (P) used to produce the desired bias using \cite{Baumann2023} approach, and the values of the parameter used in each scenario (Values)}
\label{tab:syn_ds}
\begin{tabular}{clll}
\toprule
\textbf{Label} & \textbf{Explanation} & \textbf{P} & \textbf{Values}  \\ 
\midrule
S1A & No bias present &  & \\ 
S1B & Measurement bias on $R$ & $l\_m$ & $[0.1, 0.5, 1, 1.5, 2, 3, 4, 5, 6, 7, 8, 9]$   \\ 
S1C & $R$ omitted from the dataset & $l\_o$ & True \\ 
S1D & Undersampling of group $A = 1$ & $p\_u$ & $[0.003, 0.006, 0.008, 0.01, 0.1, 0.3, 0.5]$ \\ 
S1E & Measurement bias in label $Y$ & $l\_m\_y$ & $[0.1, 0.5, 1, 1.5, 2, 3, 4, 5, 6, 7, 8, 9]$  \\ 
S1F & Conditional undersampling on $R$ & $p\_u$ & $[0.3, 0.5, 0.7, 0.9]$ \\ 
&& $l\_r$ & True \\ 
S2A & Historical bias on $R$ & $l\_h\_y$ & $[0.1, 0.5, 1, 1.5, 2, 3, 4, 5, 6, 7, 8, 9]$  \\ 
S3A & Historical bias on $Y$ & $l\_y$ & $[0.1, 0.5, 1, 1.5, 2, 3, 4, 5, 6, 7, 8, 9]$ \\ 
S4A & Historical bias on $Q$ & $l\_h\_q$ & $[0.1, 0.5, 1, 1.5, 2, 3, 4, 5, 6, 7, 8, 9]$ \\ 
\bottomrule
\end{tabular}
\end{table}

We acknowledge that the data generated have limitations. Not all kinds of bias are represented in the generated datasets; moreover, real-world data may have combined types of bias. However, this analysis allows us to identify how different biases affect differences in complexity between privileged and unprivileged groups and the relationships between complexity differences and fairness outcomes. The generated datasets can be found in the \href{https://github.com/juliettm/complexity-fairness}{GitHub} repository, which contains the code of the results of this contribution. The generation of datasets is not part of the contribution of this investigation.

\subsection{Datasets characterization}
\label{subsec:dataset_characterization}

This section provides an explanation of the characterization of the datasets used in our study, focusing on their complexity and fairness metrics. In Section \ref{subsect:complexity_datasets}, we characterize each dataset by measuring the differences in data complexity between the privileged and unprivileged groups. In Section \ref{subsect:fairness_datasets}, we assess fairness by evaluating the fairness metrics explained in Section \ref{fairness_metrics} using the solutions of different ML models for each dataset. 

\subsubsection{Complexity dataset}
\label{subsect:complexity_datasets}

Our aim is to characterize each dataset according to the differences in data complexity between privileged and unpriviledged groups. Among the range of feature-based metrics, only the F1v metric will be used. The maximum values of individual traits determine F1 and F3, which can vary between the privileged and unprivileged groups. The overlap region in F2 and the aggregated performance in F4, shaped by the order of individual feature performances, can also differ between these subgroups. 

Dimensionality metrics (T2, T3, and T4) along with the Hubs score (hubs) shall also be omitted from the analysis due to their dependency on the dimensions of the dataset. In our synthetic data, the privileged and unprivileged groups will exhibit equivalent dimensionality. The remaining metrics are part of our analysis.

\begin{table}[ht]
    \centering
    \caption{Summary of data complexity metrics used (Metric) and their short descriptions including the interpretation of results (Description)}
    \label{tab:complexitymeasuresdes}
    \begin{tabular}{>{\raggedright}m{1.5cm}>{\raggedright\arraybackslash}m{11cm}}
        \toprule
        \textbf{Metric} & \textbf{Description} \\
        \midrule
        F1v & Measures the overlap between feature values considering the optimal direction for class separation. Higher values indicate more overlap in optimal direction. \\
        \midrule
        L1 & Computes the sum of distances of misclassified examples to a linear boundary. Higher values indicate greater misclassification. \\
        \cline{1-2}
        L2 & Measures the error rate of a linear classifier. Higher values indicate more errors. \\
        \cline{1-2}
        L3 & Assesses the error rate of a linear classifier on interpolated examples. Higher values indicate more non-linear separability. \\
        \midrule
        N1 & Computes the fraction of points on the decision boundary. Higher values indicate more complex boundaries. \\
        \cline{1-2}
        N2 & Measures the ratio of intra-class to extra-class nearest neighbor distances. Higher values indicate more overlap. \\
        \cline{1-2}
        N3 & Computes the error rate of a nearest neighbor classifier. Higher values indicate more errors. \\
        \cline{1-2}
        N4 & Measures the error rate of a nearest neighbor classifier on interpolated examples. Higher values indicate more non-linear separability. \\
        \cline{1-2}
        T1 & Measures the number of hyperspheres required to cover the dataset. Higher values indicate more complex datasets. \\
        \cline{1-2}
        LSC & Computes the size of local sets around each example. A high number suggests a narrow and irregular space between classes. \\
        \midrule
        density & Measures the density of the network formed by the dataset. Lower values indicate denser networks indicating lower complexity. \\
        \cline{1-2}
        cls\_coef & Assesses the tendency of nodes to form clusters. Lower values indicate more clustering capacity. \\
        \midrule
        C1 & Measures the entropy of class proportions. Higher values indicate more balance. \\
        \cline{1-2}
        C2 & Computes the imbalance ratio for class proportions. Higher values indicate more imbalance. \\
        \bottomrule
    \end{tabular}
\end{table}

Table \ref{tab:complexitymeasuresdes} provides a description of the complexity metrics used in this paper. The mathematical definition of these metrics is provided by \cite{Lorena2019}. We modify the C1 metric so that the most challenging classification tasks correspond to the highest values by subtracting the score from 1, in \citep{Lorena2019}, the maximum C1 value of 1 is achieved for balanced problems, and subtracting the score from 1 gives the highest values for imbalanced problems. This modification enables a unified interpretation of the metric values, where higher values consistently indicate greater complexity across all metrics.

In the context of our empirical study, we partitioned each dataset into two distinct subsets based on the protected attribute (A represents the protected attribute for the generated datasets), whereby a value of 1 represents the privileged group. Subsequently, we computed an identical complexity metric for each subset. For each complexity measure, we computed the absolute difference between the complexity metric of the privileged subset and that of the unprivileged subset, as denoted in Equation \ref{eq:1}. For the sake of simplicity, we named this absolute difference (CMD) for each metric like the metric itself. Henceforth, for the remainder of this work, whenever the name of a metric is referenced, it should be understood as the absolute difference of its values for privileged and unprivileged groups.
\begin{equation}
    \label{eq:1}
    \textup{CMD} = |\textup{complexity\_metric}_{privileged} - \textup{complexity\_metric}_{unprivileged}|
\end{equation}

The results of the absolute difference in complexity metrics between privileged and unprivileged groups can be found in the supplementary material (\textit{Results-Synthetic.pdf}).

\subsubsection{Fairness dataset}
\label{subsect:fairness_datasets}

A dataset has been compiled detailing the results of fairness metrics, namely Statistical Parity (SP), Equal Opportunity (EO), and Predictive Parity (PP), for each dataset. This data includes the results of the solution of each problem using three distinct methods. For each dataset, we have solved the binary classification problem using three different machine learning methods: Logistic Regression (LR), Decision Trees (DT), and K-Nearest Neighbors (KN). 

For each method, we have used the predefined parameters except for the K-Nearest Neighbors, where we have used 10 instead of 5 neighbors. We have performed a 10-fold cross-validation and report the mean of the results for each dataset. The implementation of these solutions can be found in \citep{scikit-learn}. Neither the implementation nor the results make up part of the primary contribution of this paper; however, we present the results in the accompanying supplementary materials (\textit{Results-Synthetic.pdf}). 

The decision for choosing these methods is to have an analysis of fairness metrics with the use of different techniques in which we can evaluate how the different bias introduced can influence the fairness results in methods that operate in different ways (the linearity of the solutions, the importance of the attributes and the neighborhood of the instances).

\subsection{Finding relations between complexity differences and fairness}
\label{subsec:association_rules}

We will be using the association rules \citep{Agrawal1994} to discover patterns that relate differences in complexity metrics for privileged and unprivileged groups to the fairness metrics calculated for the solution of each problem. 

The primary objective of association rule mining is to identify strong rules that describe how the presence of certain items in a dataset implies the presence of other items; in our case, we will identify the presence of difference in complexity metrics implying the presence of unfairness. Specifically, we are going to identify patterns that relate differences in complexity higher than $0.1$ with unfairness, i.e. values in the metrics outside the fair interval $[-0.1, 0.1]$.

An association rule is typically represented in the form $Antecedent \rightarrow Consequent$, where $Antecedent$ and $Consequent$ are sets of items (itemsets). The strength of these rules is evaluated on the basis of different metrics, let's consider an association rule  $X \rightarrow Y$.

\begin{itemize}
    \item \textbf{Support}: Measures the proportion of transactions in the dataset that contain both $X$ and $Y$.
    \begin{equation}
        Support = \frac{Frequency(X,Y)}{N}
    \end{equation}
    where $Frequency(X,Y)$ is the number of appearances of both $X$ and $Y$ and $N$ the number of rows of the dataset (transactions).
    
    \item \textbf{Confidence}: Measures the likelihood of finding the consequence $Y$ in transactions under the condition that these transactions also contain the antecedent $X$.
    \begin{equation}
        Confidence = \frac{Support(X,Y)}{Support(X)}
    \end{equation}
        
    \item \textbf{Lift}: Lift assesses how much more likely the consequent $Y$ is to occur with the antecedent $X$ than if $Y$ were independent of $X$. A lift value greater than 1 indicates a positive correlation between $X$ and $Y$, meaning that $Y$ is more likely to occur when $X$ is present. This measures the strength of the association between X and Y.
    \begin{equation}
        Lift = \frac{Confidence(X,Y)}{Support(Y)}
    \end{equation}
\end{itemize}

The mining association rules process encompasses two principal steps. In our approach to rule generation, we employ the mlextend library \citep{raschkas_2018_mlxtend}. Initially, we used the Apriori function to identify frequent itemsets, adhering to a minimum support specified, as outlined in $0.1$. Thereafter, we proceed to generate rules, focusing on those that exhibit a minimum lift value of 1. The results obtained will allow us to derive relationships between the differences in complexity between privileged and unprivileged subgroups and fairness.

\section{Analysis and Results}
\label{lab:results}

This section presents a comprehensive examination of our experimental findings. We begin by comparing different bias-generating mechanisms to understand how they affect subgroup complexity differences; addressing the first investigation question on intrinsic complexity differences in Section \ref{subsect_complexity}. Next, we evaluate the results of the fairness metrics obtained for each bias scenario in Section \ref{subsect:anlysis_fairness}. We assess whether complexity differences can act as indicators of unfairness (Section \ref{subsect:assoc_rules}), thus guiding early-stage data corrections and model adjustments to address the second investigation question. 

\subsection{Analysis of differences in complexity}
\label{subsect_complexity}

\begin{figure}[ht]
    \centering
    \includegraphics[width=0.98\linewidth]{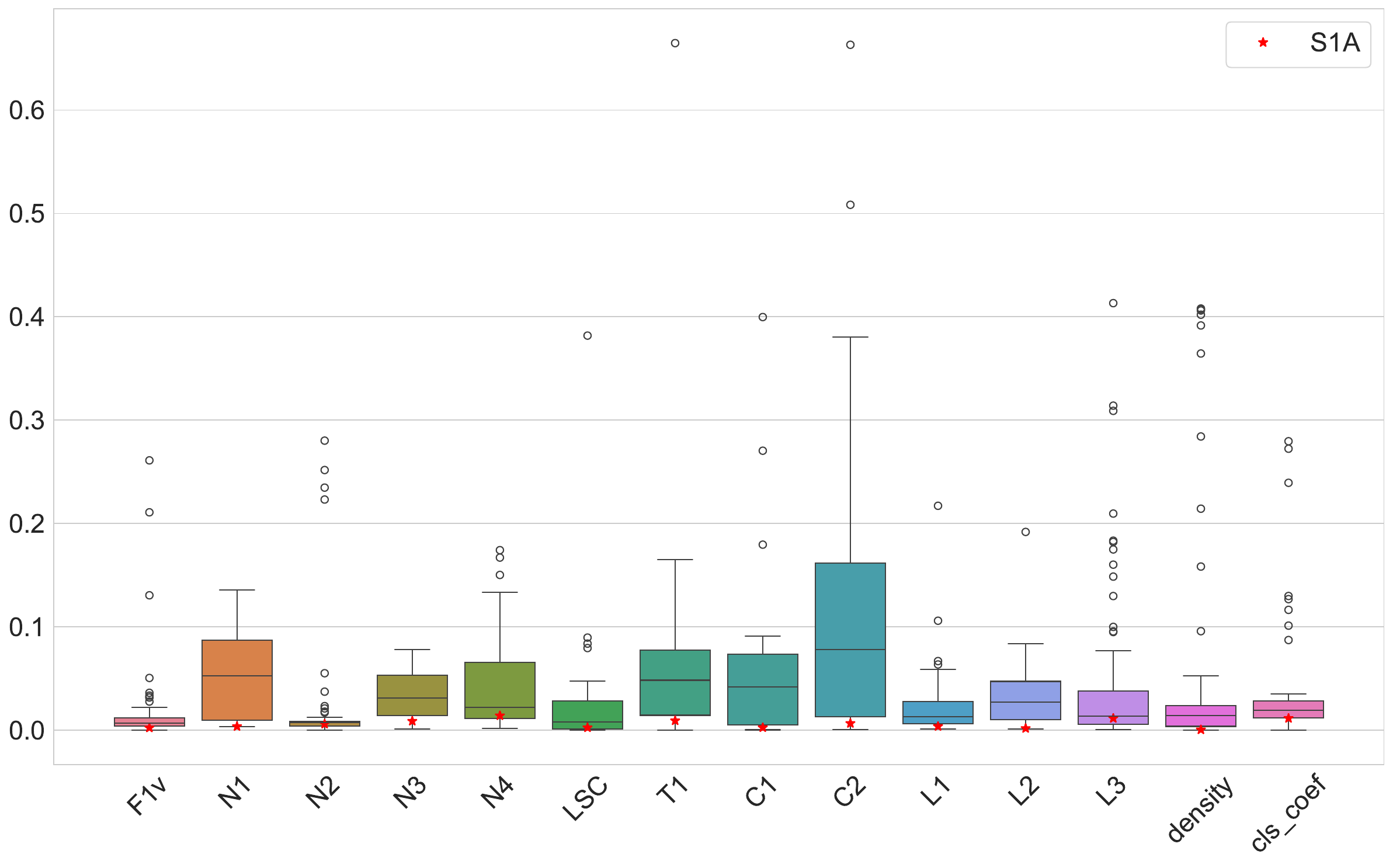}
    \caption{Distribution of absolute differences in various complexity metrics when comparing privileged and unprivileged groups. Each metric (labeled along the horizontal axis) is represented by a separate distribution illustrating how its absolute difference values are spread. The vertical axis reflects the magnitude of these differences. The unbiased dataset marked as a red star. Each complexity metric absolute difference exhibits unique range coverage and variability patterns. }
    \label{fig:all-bp}
\end{figure}

\paragraph{Complexity metrics distribution} The visualization of the box plot in Figure \ref{fig:all-bp} reveals different distribution patterns between the differences in complexity metrics for the privileged and unprivileged groups. Each complexity metric difference exhibits unique range coverage and variability patterns that require a detailed examination. However, as observable, the unbiased dataset marked as a red star (Experiment S1A) consistently takes minimum values for all the differences, which illustrates that the variation in bias introduced is increasing the differences in complexity metrics between the groups in different magnitudes.

Importantly, none of the differences in complexity metrics extend throughout their entire theoretical range of [0,1]. This implies that our generated datasets do not cover the entire array of complexity-difference scenarios, which highlights the limitation of not capturing the complete spectrum of potential complexity situations.

\paragraph{Complexity metrics distribution by bias scenarios}

\begin{figure}[ht]
    \centering
    \includegraphics[width=0.78\linewidth]{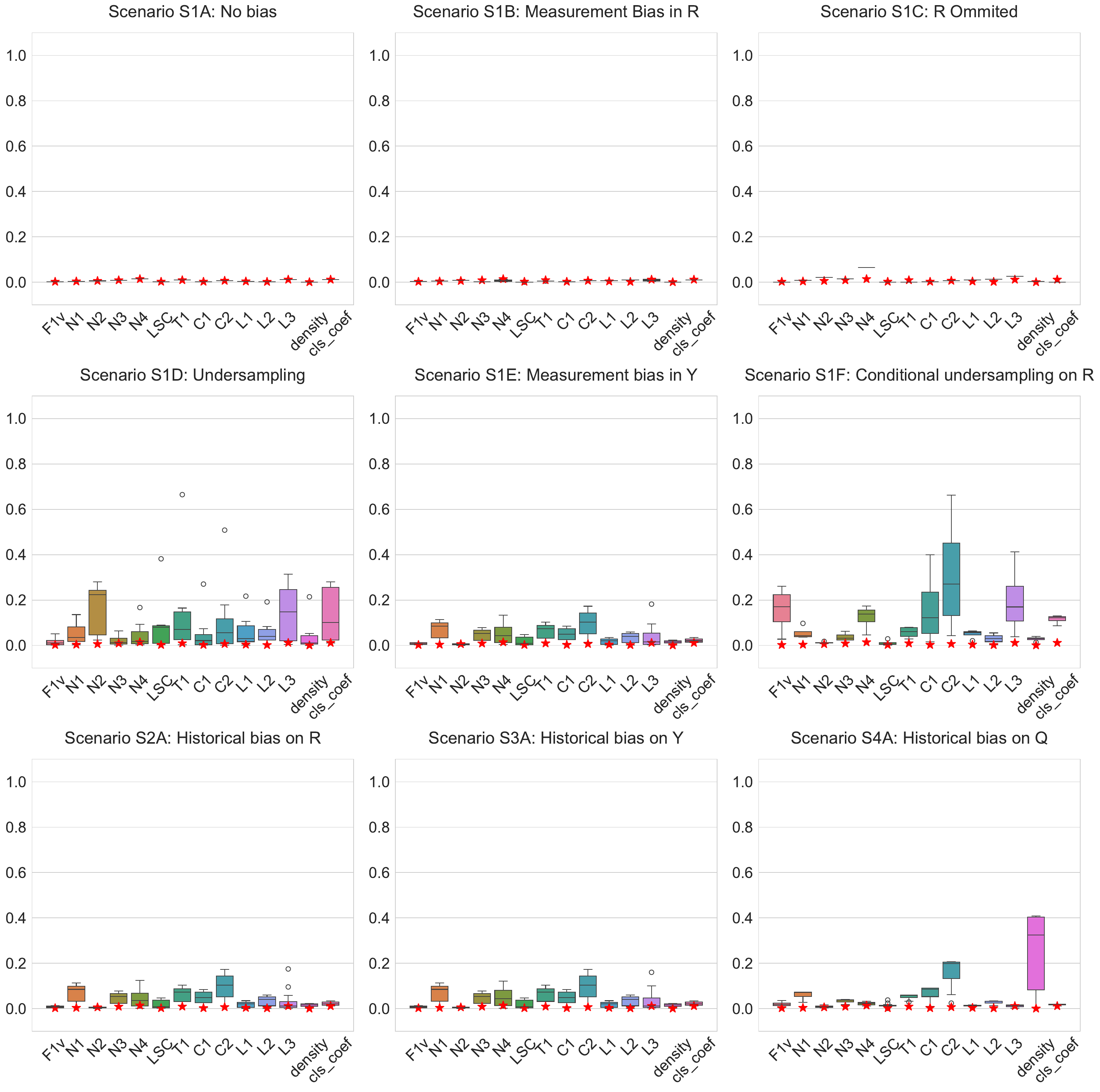}
    \caption{Distribution of complexity metrics absolute differences by each bias scenario. Each subplot shows boxplot distributions for a specific bias scenario, with red stars indicating the baseline (S1A) values. S1B and S1C show no differences in complexity. S1E, S2A and S3A show similar differences distributions and the rest of the scenarios present different unique distributions of the complexity metrics.
    }
    \label{fig:complexity-scenarios}
\end{figure}

To support the analysis of differences in the ranges of the metrics obtained, we have plotted the distribution of each complexity metric by scenario compared to the baseline scenario, which can be seen in Figure \ref{fig:complexity-scenarios}, where each subplot represents a different scenario from S1A to S4A. We can identify that the absolute differences between privileged and unprivileged values for all complexity metrics are almost imperceptible for scenarios S1A, S1B, and S1C. We can discern similarities between the distributions of the metrics observed in S1E and S3A where bias is inflicted on Y (measurement and historical, respectively) and also with the scenario S2A (historical bias on R). 

\begin{figure}[ht]
    \centering
    \includegraphics[width=0.78\linewidth]{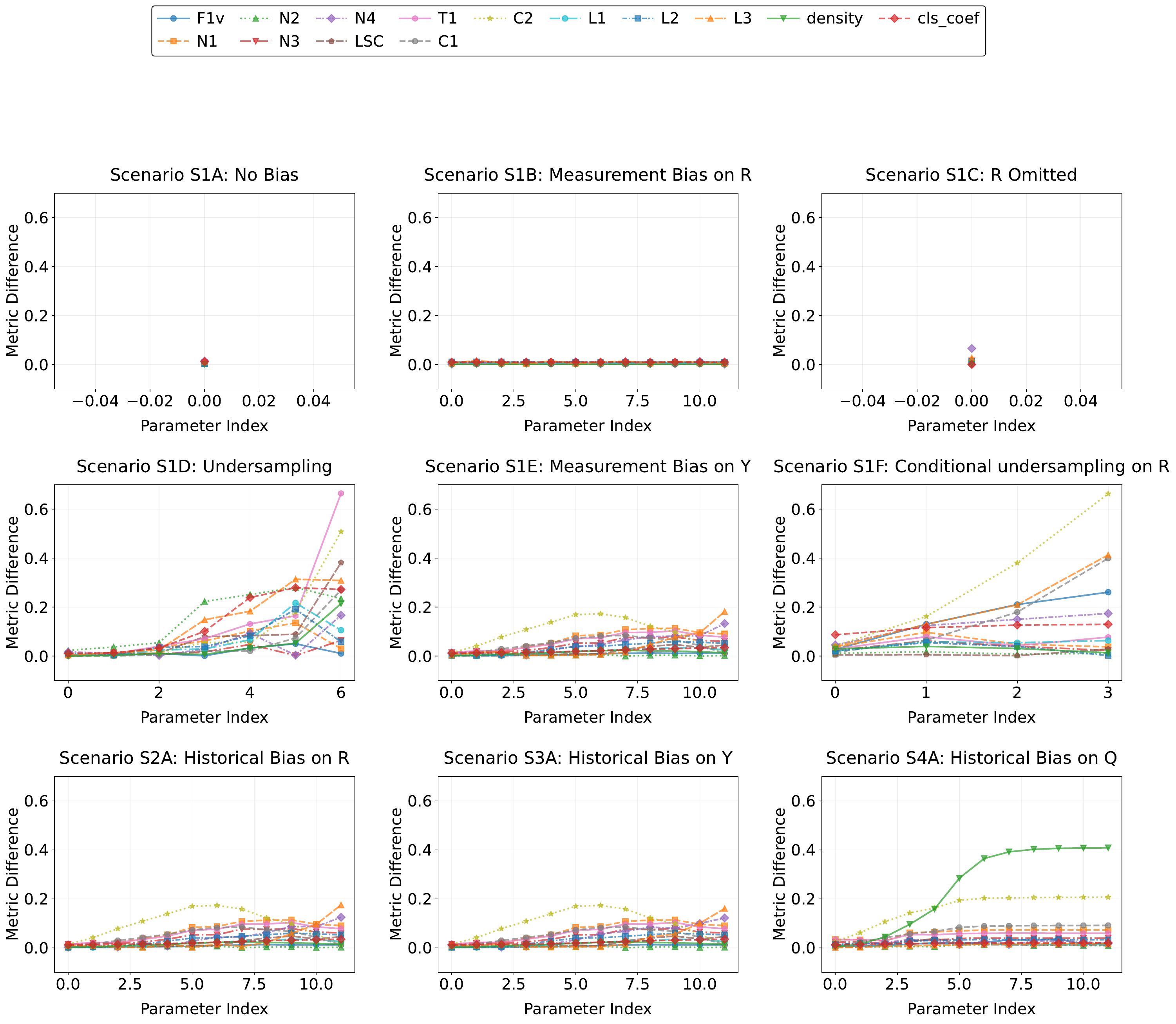}
    \caption{Evolution of complexity metrics difference by bias parameter. Each subplot represents a bias scenario in which each line represent a complexity metric. The horizontal axis represents the index of the array of values (Values column on Table \ref{tab:syn_ds}) of the parameter (P value on Table \ref{tab:syn_ds}) controlling the bias magnitude on each scenario. The vertical axes represent the magnitude of the differences in complexity. The evolution of complexity metrics across bias parameters reveals that while some metrics consistently increase with bias, others fluctuate based on specific scenario characteristics.}
    \label{fig:complexity-by-scenario}
\end{figure}

As the parameters used to generate various bias scenarios are increased, the complexity metric does not invariably exhibit a consistent increase or decrease in different scenarios (see Figure \ref{fig:complexity-by-scenario}). This can be observed in the metric C2 for Scenarios S1E, S2A, and S3A. However, there is a tendency to increase for most differences, especially for scenarios S1D, S1F, and S4A. 

This plot also confirms the minimum incidence in the differences between groups for all the metrics in scenarios S1A, S1B, and S1C. This suggests that while the differences in the non-biased scenario for all the metrics are near zero supporting our theory there are scenarios when measurement bias or omitted variable bias can be present but the differences in complexity between groups are insignificant.

In addition, we constructed a comparative analysis of the variability of each complexity metric difference across different scenarios, plotted by metric (see Figure~\ref{fig:metrics-scenario}). This comparison reveals which complexity metrics are the most sensitive to different types of bias and which remain stable regardless of the bias introduced. The analysis reveals several notable patterns:

\begin{itemize}
    \item \textbf{Feature-based metrics (F1v):} F1v show differences in the scenario S1F (Conditional Undersampling on R).
    
    \item \textbf{Neighborhood-based metrics (N1-N4, LSC and T1):} N1, N3, N4, LSC, and T1 show differences for all scenarios except by S1A, S1B and S1C. It is notable that S1D (Undersampling) affects LSC and T1 the most. The differences for the N2 are near zero except for S1D.

    \item \textbf{Class imbalance metrics (C1, C2)}: C1 exhibits medium differences in complexity values that are noticeable for S1F (Conditional Undersampling on R). On the other hand, C2 shows more pronounced differences variations specially for S1D, S1F and S4A (Undersampling, Conditional Undersampling on R and Historical Bias on Q).

    \item \textbf{Linearity metrics (L1-L3)}: L1 and L2 do not show very high differences except for the S1D values. L3 show higher differences in the differences mostly for S1D and S1F (Undersampling and Conditional Undersampling).

    \item \textbf{Network metrics (density, clustering coefficient)}: Density and clustering coefficient show minimal differences for almost all scenarios. While density show high differences mostly for S4A (Historical Bias on Q), clustering coefficient show the biggest differences for S1D (Undersampling).

\begin{figure}[ht]
    \centering
    \includegraphics[width=0.78\linewidth]{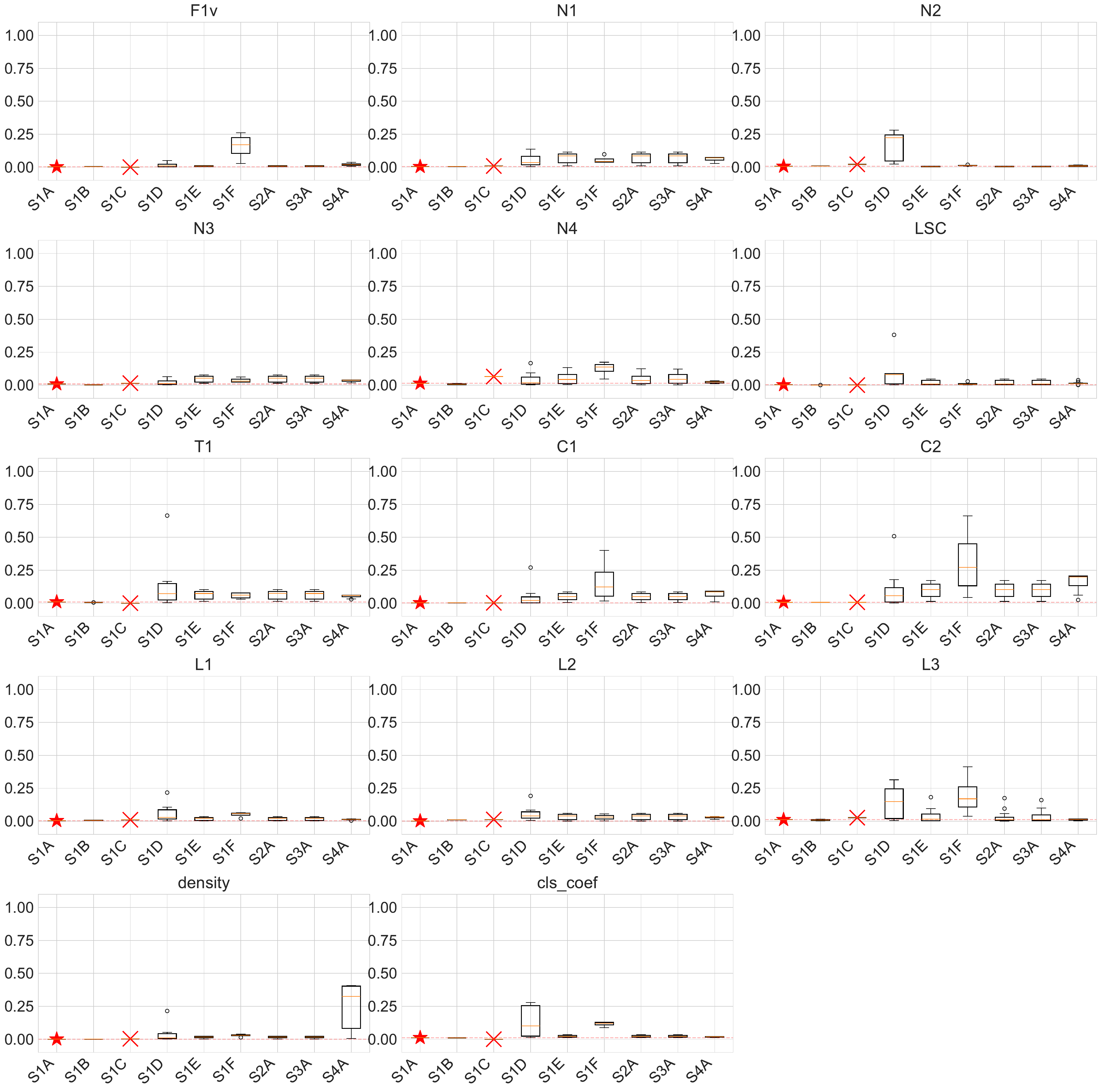}
    \caption{Distribution of complexity metric values across bias scenarios. Each subplot shows how a specific complexity metric varies in different bias scenarios, with each scenario labeled on the horizontal axis represented by a separate distribution illustrating how its absolute difference values spread. The vertical axis reflects the magnitude of these differences.}
    \label{fig:metrics-scenario}
\end{figure}
\end{itemize}

Our analysis of complexity metrics differences between privileged and unprivileged groups highlights key patterns in how bias affects data complexity. The results indicate that the unbiased dataset (S1A) consistently exhibits minimal complexity differences, supporting the hypothesis that biases in datasets lead to disparities between groups. However, not all complexity metrics respond uniformly to bias introduction, suggesting that different forms of bias influence the complexity of the dataset in different ways.

Examining the bias scenarios, we observe that the measurement bias in Y (S1E) and the historical bias in R and Y show a similar distribution of metric differences. We have also analyzed how different types of bias affect complexity metrics in different ways, indicating varying reliability in detecting bias. Furthermore, the evolution of complexity metrics across bias parameters (Figure \ref{fig:complexity-by-scenario}) reveals that while some metrics consistently increase with bias, others fluctuate based on specific scenario characteristics.

\begin{figure}
    \centering
    \includegraphics[width=0.6\linewidth]{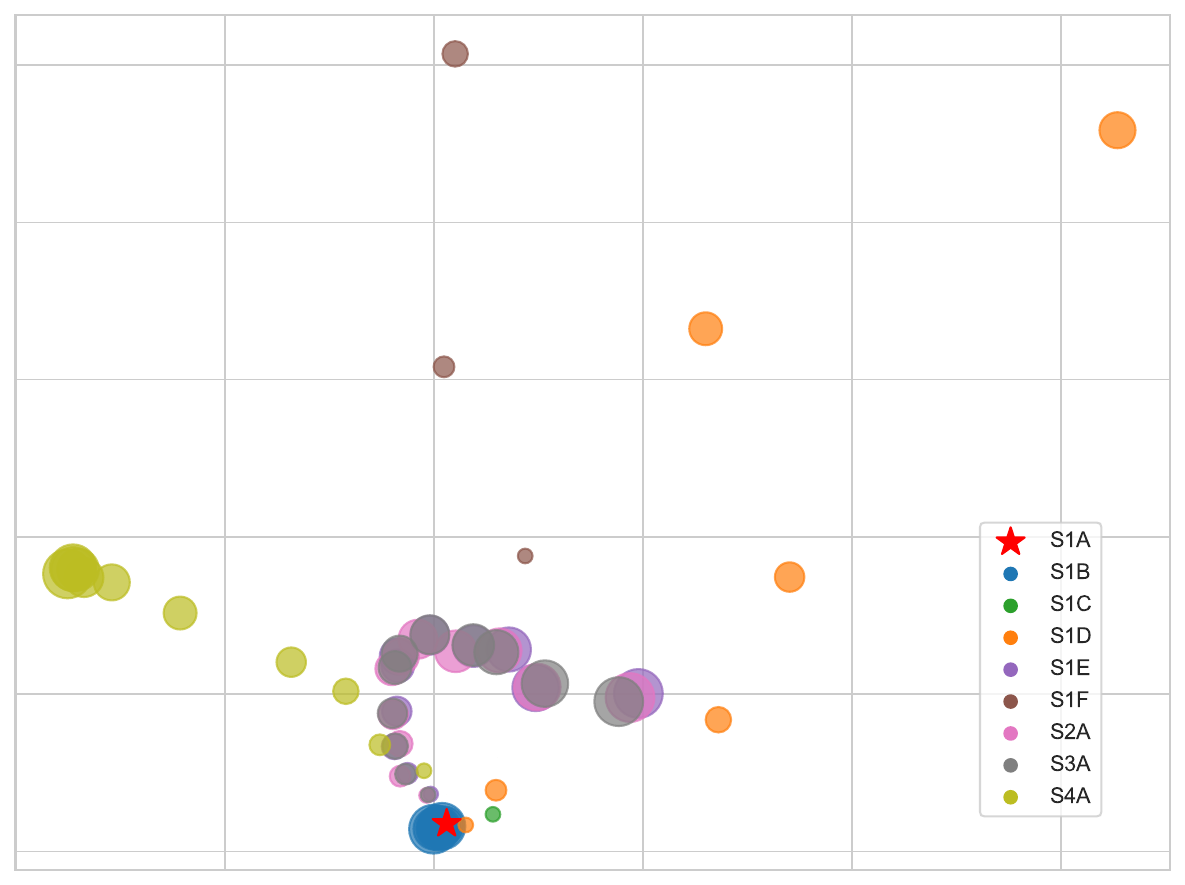}
    \caption{Two-dimensional MDS visualization of synthetic datasets complexity metrics absolute differences colored by bias scenario. The baseline scenario (S1A) is marked with a star. Point sizes correspond to bias parameter magnitude, with larger points indicating stronger bias. Datasets that appear close to each other in the plot are more similar, while datasets that appear far apart are more dissimilar. S1D, S1F and S4A show the biggest differences. S1E, S2A and S3 show almost identical distributions.}

    \label{fig:mds-all-datasets}
\end{figure}

\paragraph{Characterization of the datasets using complexity metrics - Dimensional Reduction Analysis}
We transformed the datasets of the differences in complexity metrics using multidimensional scaling (MDS) \citep{scikit-learn}. The visualization of the two dimensions resulting that represent the characteristic of the datasets reveals clustering patterns and relationships between the datasets with different types of bias, where proximity in the two-dimensional space indicates similarity in complexity characteristics (see Figure \ref{fig:mds-all-datasets}). The baseline scenario (S1A), marked with a star, serves as a reference point to compare the impact of different types of bias. The interpoint distances are computed using the Euclidean metric.

The varying point sizes, representing the magnitude of bias parameters, demonstrate a clear pattern where datasets with higher bias parameter values tend to deviate further from the baseline scenario (S1A). MDS coordinates were computed using the complete set of complexity metrics, providing a general view of how different bias types affect the overall characteristics of the dataset.
    
Measurement bias scenarios (S1B, S1E) form distinct patterns in the visualization, suggesting these bias types do not induce consistent patterns of complexities differences, S1B values are very close to the non-biased scenario while S1E datastes spread together with S2A and S3A sugesting the same similarities observed in the complexity differences box plot distribution. The omitted variable bias (S1C) is close to the non-biased scenario, while S1D, S1F and S4A (Undersampling and Historical bias on Q) show more dispersion from the non-biased scenario S2A. This dimensional reduction analysis suggests that different types of bias induce distinctive patterns in the overall complexity characteristics of datasets leading to differences between privileged and unprivileged group differences, with some types of bias showing more consistent and pronounced effects than others.

It is important to note that these patterns are observed within our specific experimental setup and dataset characteristics. The differences in complexity metrics do not span their full theoretical ranges, suggesting potential limitations in capturing all cases. In addition, some bias scenarios show variable effects, indicating that complexity patterns alone may not be sufficient for the identification of bias types.

\subsection{Analysis of Fairness Results}
\label{subsect:anlysis_fairness}

\begin{figure}[ht]
    \centering
    \includegraphics[width=0.7\linewidth]{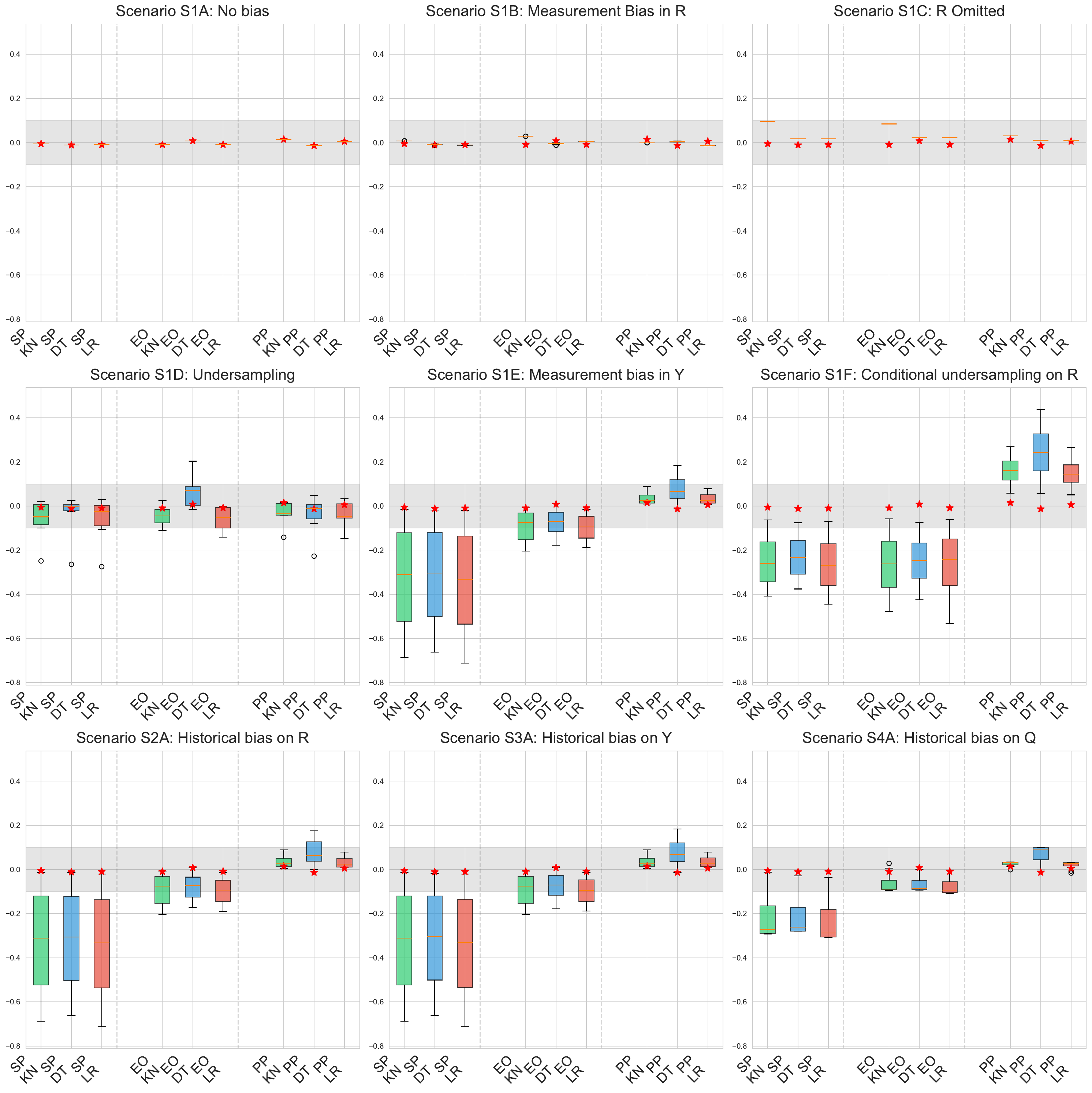}
    \caption{Distribution of fairness metrics results (Statistical Parity (SP), Equal Opportunity (EO), and Predictive Parity (PP)) computed using different methods (k-Nearest Neighbors (KN), Decision Trees (DT), and Logistic Regression (LR)) by each bias scenario. The horizontal axis show fairness metrics results for different ML methods represented by a separate distribution illustrating how its values spread. The vertical axis reflects the magnitude of fairness metrics. The gray band representing acceptable fairness bounds (±0.1). S1A, S1B and S1C are confirmed to have fair results. The biggest differences can be observed in SP in the rest of the scenarios.}
    \label{fig:fairness-distribution}
\end{figure}

Proceeding with the analysis of the effect on fairness of the different bias introduced in the data, Figure~\ref{fig:fairness-distribution} shows the fairness metrics distribution. Each subplot illustrates a scenario, and within each subplot, the variability of the fairness metrics (SP, EO and PP) across different methods (KN, LR and DT) is represented using boxplots.

\begin{itemize}
    \item \textbf{Baseline Scenario (S1A):} All fairness metrics are located around zero for the three algorithms indicating fair predictions in the absence of bias.

    \item \textbf{Measurement bias in R (S1B) and Omitted variable bias in R (S1C) Scenarios}: Show minimal variations, and the values remain fair.

    \item \textbf{Representation Bias Impact (S1D, S1F) Scenarios:}
    \begin{itemize}
        \item S1D show moderate variation of the fairness metrics for all algorithms but mantaining almost all values in the fair bounds. 
        \item S1F shows the most severe fairness violations mostly in  PP with a consistent pattern of positive bias in PP metrics across all algorithms while SP and EO metrics show negative deviations.
    \end{itemize}
    
    \item \textbf{Measurement Bias in Y Scenario (S1E):} shows substantial negative deviations in SP metrics across all algorithms. EO and PP metrics demonstrate increasing disparity, particularly for DT. KN and LR exhibit similar patterns of fairness degradation.
    
    \item \textbf{Historical Bias Scenarios (S2A, S3A, S4A):} Strong negative impact on SP metrics (-0.5) for all algorithms. EO metrics show moderate variation but remain within acceptable bounds.  PP metrics demonstrate positive bias, particularly for DT.
\end{itemize}

It is important to note that the distributions in various metrics and algorithms demonstrate substantial similarities within scenarios S1E, S2A, and S3A. Upon examining algorithm-specific patterns, the most discernible differences emerge in the application of Decision Trees (DT) as opposed to Logistic Regression (LR) and k-Nearest Neighbors (KN), with DT presenting reduced discrepancies in SP but more pronounced variances in PP.

These observations reveal that different types of bias affect fairness metrics in different ways, with some combinations of bias type and algorithm producing more severe fairness violations than others. The analysis suggests that representation bias (S1F) has the most pronounced effect on Predictive Parity, but also affects Statistical Parity and Equal Opportunity. Measurement bias (S1E) and historical bias scenarios (S2A, S3A and S4A) primarily impact Statistical Parity. Remarkably, the choice of an algorithm can influence the magnitude of fairness violations under biased conditions.

\begin{figure}[p!]
    \centering
    \includegraphics[width=0.8\linewidth]{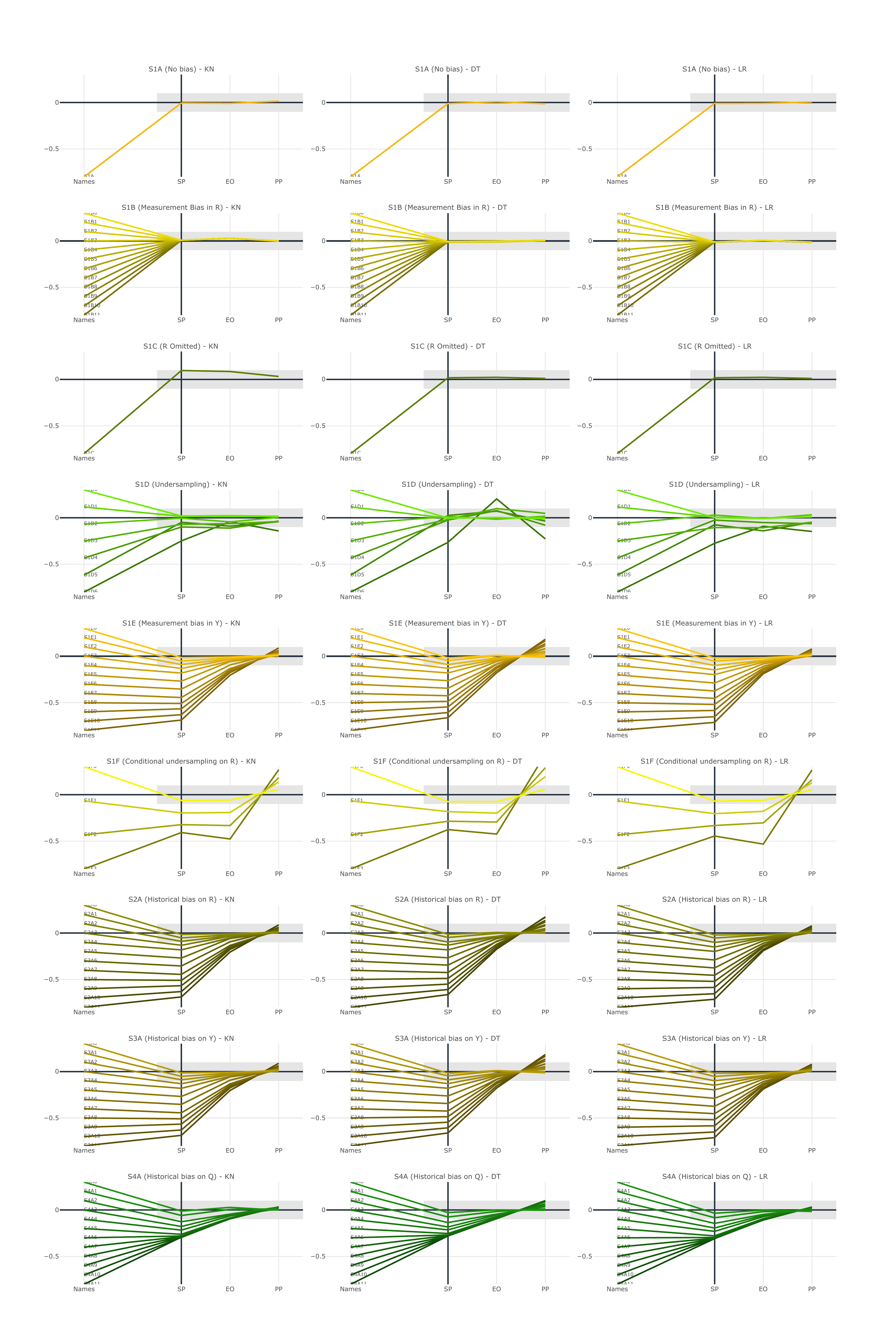}
    \caption{Parallel coordinates visualization of fairness metrics across different bias scenarios and algorithms. Each dataset is represented by a line in the subplots. S1A, S1B and S1C are confirmed to have fair results. The lines show consistent trends on negative results for SP and EO and positive values for PP.}
    \label{fig:fairness-paralell-coordinates}
\end{figure}

The visualization of parallel coordinates (see Figure \ref{fig:fairness-paralell-coordinates}) effectively reveals the relationships between different fairness metrics in various bias scenarios and algorithms. Each vertical axis represents one of three fairness metrics (SP, EO, and PP), while each line traversing these axes shows how a specific parameter configuration performs across all metrics. 

The plot arrangement in Figure \ref{fig:fairness-paralell-coordinates} allows for direct comparison between algorithms (across columns) and bias scenarios (across rows), highlighting how different types of bias affect the trade-offs between fairness metrics. Notable patterns include consistent negative SP values in historical bias scenarios (S2A-S4A), significant metric variations in representation bias cases (S1F).

In scenarios S1A, S1B, and S1C, the metrics exhibit near zero values. These scenarios similarly demonstrate the absence of disparity in complexity metrics between the privileged and unprivileged groups. This observation corroborates the hypothesis that the disparities between the groups may be a hint of unfairness in the solutions.

\subsection{Analysis of association rules to discover relationships between complexity differences and fairness}
\label{subsect:assoc_rules}

\begin{figure}[h!]
    \centering
    \includegraphics[width=0.6\linewidth]{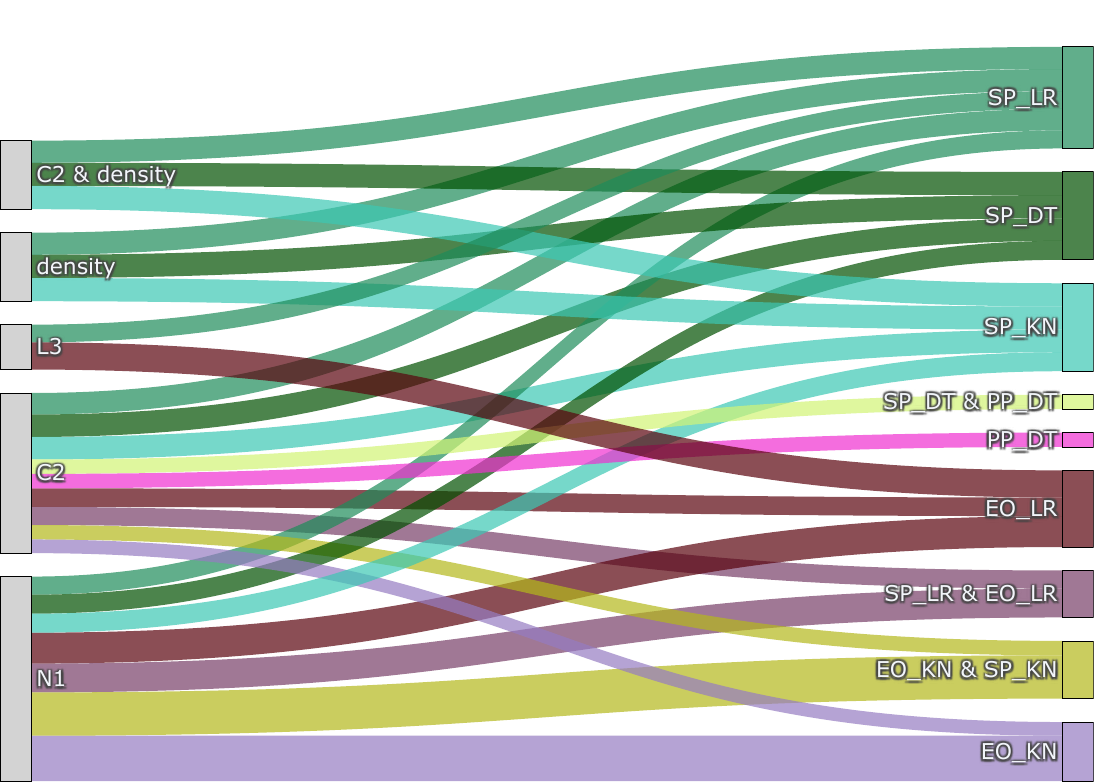}
    \caption{Association rules. The left side is the antecedents of the rules (complexity metrics differences) and the right side represents fairness metrics, consequent of the rules. The squared lift values are used as the width of the links between the rules to emphasize the differences. N1 and C2 differences appear frequently as antecedents making them indicators of different fairness measures.}
    \label{fig:association_rules}
\end{figure}

After applying the process of generating rules, based on the complexity differences and fairness metrics results, a total of 24 association rules have been derived. The rules have a minimum support threshold of $0.1$ and a lift exceeding $1$, and they have complexity metric differences as antecedents and a fairness metric as a consequent. Table \ref{tab:association_rules} shows the generated rules as well as the values for support, confidence and lift while Figure \ref{fig:association_rules} shows a visualization of the rules in a Sankey diagram.

\begin{table}[ht]
    \centering
    \caption{association rules derived from complexity metrics differences and fairness metrics in the synthetic datasets. The table show the values of the antecedents and consequent, as well as the support of each one (Sup\_A and Sup\_C) followed by the support (Sup), Confidence and Lift of each rule.}
    \label{tab:association_rules}

    \begin{tabular}{llrrrrr}
        \toprule
        Antecedent & Consequent & Sup\_A & Sup\_C & Sup & Confidence & Lift \\
        \midrule

        C2 & SP\_DT & 45\% & 56\% & 44\% & 0.97 & 1.73 \\
        C2 & SP\_LR & 45\% & 59\% & 44\% & 0.97 & 1.65 \\
        C2 & SP\_KN & 45\% & 56\% & 44\% & 0.97 & 1.73 \\
        C2 & EO\_LR & 45\% & 38\% & 25\% & 0.55 & 1.42 \\
        C2 & SP\_LR \& EO\_LR & 45\% & 37\% & 23\% & 0.52 & 1.39 \\
        N1 & EO\_KN & 15\% & 26\% & 14\% & 0.91 & 3.49 \\
        N1 & EO\_LR & 15\% & 38\% & 14\% & 0.91 & 2.37 \\
        C2 \& density & SP\_LR & 12\% & 59\% & 12\% & 1.00 & 1.70 \\
        N1 & SP\_LR \& EO\_LR & 15\% & 37\% & 12\% & 0.82 & 2.21 \\
        density & SP\_LR & 12\% & 59\% & 12\% & 1.00 & 1.70 \\
        N1 & SP\_LR & 15\% & 59\% & 12\% & 0.82 & 1.39 \\
        C2 & EO\_KN \& SP\_KN & 45\% & 25\% & 12\% & 0.27 & 1.11 \\
        N1 & SP\_DT & 15\% & 56\% & 12\% & 0.82 & 1.46 \\
        N1 & EO\_KN \& SP\_KN & 15\% & 25\% & 12\% & 0.82 & 3.32 \\
        density & SP\_KN & 12\% & 56\% & 12\% & 1.00 & 1.78 \\
        C2 & EO\_KN & 45\% & 26\% & 12\% & 0.27 & 1.05 \\
        N1 & SP\_KN & 15\% & 56\% & 12\% & 0.82 & 1.46 \\
        C2 & SP\_DT \& PP\_DT & 45\% & 25\% & 12\% & 0.27 & 1.11 \\
        C2 \& density & SP\_DT & 12\% & 56\% & 12\% & 1.00 & 1.78 \\
        density & SP\_DT & 12\% & 56\% & 12\% & 1.00 & 1.78 \\
        C2 & PP\_DT & 45\% & 25\% & 12\% & 0.27 & 1.11 \\
        C2 \& density & SP\_KN & 12\% & 56\% & 12\% & 1.00 & 1.78 \\
        L3 & SP\_LR & 14\% & 59\% & 11\% & 0.80 & 1.36 \\
        L3 & EO\_LR & 14\% & 38\% & 11\% & 0.80 & 2.09 \\

        \bottomrule
        \end{tabular}
    
\end{table}

Differences in class imbalance between the privileged and unprivileged groups (C2) are consistently associated with fairness disparities for all consequent, which implies that addressing class imbalance could be a key focus area to improve fairness. This indicates that when class distributions are imbalanced, it tends to affect fairness metrics, possibly leading to disparities in outcomes across different groups. The strongest association is with SP for the three methods (DT, KN, and LR), suggesting that the imbalance is most strongly correlated with differences in Statistical Parity. 

Differences in the decision boundary cases (N1) show significant associations with SP and EO for the KN and LR algorithm. The average density of network differences (density) also shows an association with SP for the KN algorithm.

Non-linearity in a linear classifier (L3) is correlated with the Statistical Parity (SP) and Equal Opportunity (EO) metrics of the outcomes of Logistic Regression (LR). Although these results may be straightforward, the extracted rules make their clarity more apparent. In addition, the differences in densities (density) and their combination with the imbalance ratio (C2) are related to the unfairness measured as SP in all the techniques applied.

These association rules confirm meaningful, quantifiable links between subgroup complexity differences and fairness outcomes (addressing our second question). In particular, class imbalance (C2) and boundary overlap metrics (N1) differences frequently signal where Statistical Parity (SP) and Equal Opportunity (EO) will degrade, demonstrating a clear pathway for proactive bias detection. Moreover, the consistent appearance of these complexity factors as antecedents (along with density and non-linearity metrics) highlights how group-specific complexity can reliably indicate when algorithmic decisions are likely to become unfair.

\section{Application in real-world problems}
\label{lab:real_world}

Using sources cited in \cite{Fabris2022} cataloging diverse datasets for the analysis of fairness in different machine learning problems and other work on classification, we accumulated a total of 30 datasets for classification tasks. Table \ref{tab:realdatasets} displays the datasets included in this research, along with the output name for each problem and its favorable label, the protected attribute, and its privileged value. The table also indicates the number of rows and attributes for each dataset. We do not provide the data in our associated material, as each dataset is subject to diverse rights, but Table \ref{tab:realdatasets} specifically references the origin of the data.

These datasets represent different classification tasks, each of which is assumed to have one binary protected attribute (PA column). For our experiments, we preprocessed the datasets to uniformly assign the value $1$ to the favorable label of the output and the privilege value of the protected attribute (preprocessing steps can be found in \href{https://github.com/juliettm/complexity-fairness}{GitHub}).

For each of them, we have computed the differences in all the complexity metrics previously analyzed, and the fairness metrics SP, EO, and PP values using the results of KN, DT, and LR method. The results are listed in the complementary material (\textit{Results-Real-World.pdf}).

Table \ref{tab:real-world-rules} presents the evaluation of the association rules obtained in Subsection \ref{subsect:assoc_rules} for the 30 real-world problems. 

\begin{table}[ht]
    \centering
    \caption{Results of the rules applied to real-world problems. The table shows each rule antecedent and consequent and the values of the support of each one (Sup\_A and Sup\_C) followed by the support (Sup), confidence, and lift computed using the results in real-world problems.}
    \label{tab:real-world-rules}
\begin{tabular}{llrrrrr}
\toprule
Antecedent & Consequent & Sup\_A & Sup\_C & Sup & Confidence & Lift \\
\midrule
    C2 & SP\_DT & 60\% & 47\% & 40\% & 0.67 & 1.43 \\
    C2 & SP\_LR & 60\% & 67\% & 53\% & 0.89 & 1.33 \\
    C2 & SP\_KN & 60\% & 47\% & 40\% & 0.67 & 1.43 \\
    C2 & EO\_LR & 60\% & 50\% & 40\% & 0.67 & 1.33 \\
    C2 & SP\_LR \& EO\_LR & 60\% & 50\% & 40\% & 0.67 & 1.33 \\
    N1 & EO\_KN & 30\% & 40\% & 17\% & 0.56 & 1.39 \\
    N1 & EO\_LR & 30\% & 50\% & 20\% & 0.67 & 1.33 \\
    C2 \& density & SP\_LR & 10\% & 67\% & 10\% & 1.00 & 1.50 \\
    N1 & SP\_LR \& EO\_LR & 30\% & 50\% & 20\% & 0.67 & 1.33 \\
    density & SP\_LR & 10\% & 67\% & 10\% & 1.00 & 1.50 \\
    N1 & SP\_LR & 30\% & 67\% & 30\% & 1.00 & 1.50 \\
    C2 & SP\_KN \& EO\_KN & 60\% & 40\% & 33\% & 0.56 & 1.39 \\
    N1 & SP\_DT & 30\% & 47\% & 27\% & 0.89 & 1.90 \\
    N1 & SP\_KN \& EO\_KN & 30\% & 40\% & 17\% & 0.56 & 1.39 \\
    density & SP\_KN & 10\% & 47\% & 10\% & 1.00 & 2.14 \\
    C2 & EO\_KN & 60\% & 40\% & 33\% & 0.56 & 1.39 \\
    N1 & SP\_KN & 30\% & 47\% & 17\% & 0.56 & 1.19 \\
    C2 & SP\_DT \& PP\_DT & 60\% & 43\% & 37\% & 0.61 & 1.41 \\
    density \& C2 & SP\_DT & 10\% & 47\% & 7\% & 0.67 & 1.43 \\
    density & SP\_DT & 10\% & 47\% & 7\% & 0.67 & 1.43 \\
    C2 & PP\_DT & 60\% & 57\% & 47\% & 0.78 & 1.37 \\
    density \& C2 & SP\_KN & 10\% & 47\% & 10\% & 1.00 & 2.14 \\
    L3 & SP\_LR & 40\% & 67\% & 30\% & 0.75 & 1.12 \\
    L3 & EO\_LR & 40\% & 50\% & 27\% & 0.67 & 1.33 \\
    \bottomrule
    \end{tabular}

\end{table}

Notably, the highest lift value ($2.14$) is observed in rules where density is the antecedent, particularly influencing SP in K-Nearest Neighbors (KN), suggesting a strong dependency between local instance distribution and fairness disparities. Similarly, rules involving C2 (class imbalance) frequently exhibit high confidence. Some rules where N1 (decision boundary complexity) is the antecedent display moderate confidence ($0.56$) and lower lift values ($1.19, 1.39$), signifying weaker associations with fairness outcomes. The variability in confidence and lift across different metrics highlights the non-uniform influence of complexity on fairness, reinforcing the need for classifier-specific fairness interventions.

These findings show that the same patterns observed in synthetic data (particularly regarding class imbalance, density, and boundary overlap) also emerge in diverse real-world problems, although with additional complexity and overlapping bias sources. Rules with high lift and confidence underscore how localized data characteristics differences (e.g., subgroup density, imbalance) consistently drive fairness disparities across multiple algorithms. Meanwhile, variations in confidence and lift illustrate that no single complexity metric universally predicts unfairness; each classifier and dataset combination may exhibit unique susceptibility to certain complexity indicators. Hence, monitoring these subgroup-specific complexity signals early in a real-world pipeline can flag data-driven bias risks, prompting more targeted interventions, like rebalancing or refining local decision boundaries, to enhance fairness in practical ML deployments.

\begin{figure}[h!]
    \centering
    \includegraphics[width=0.7\linewidth]{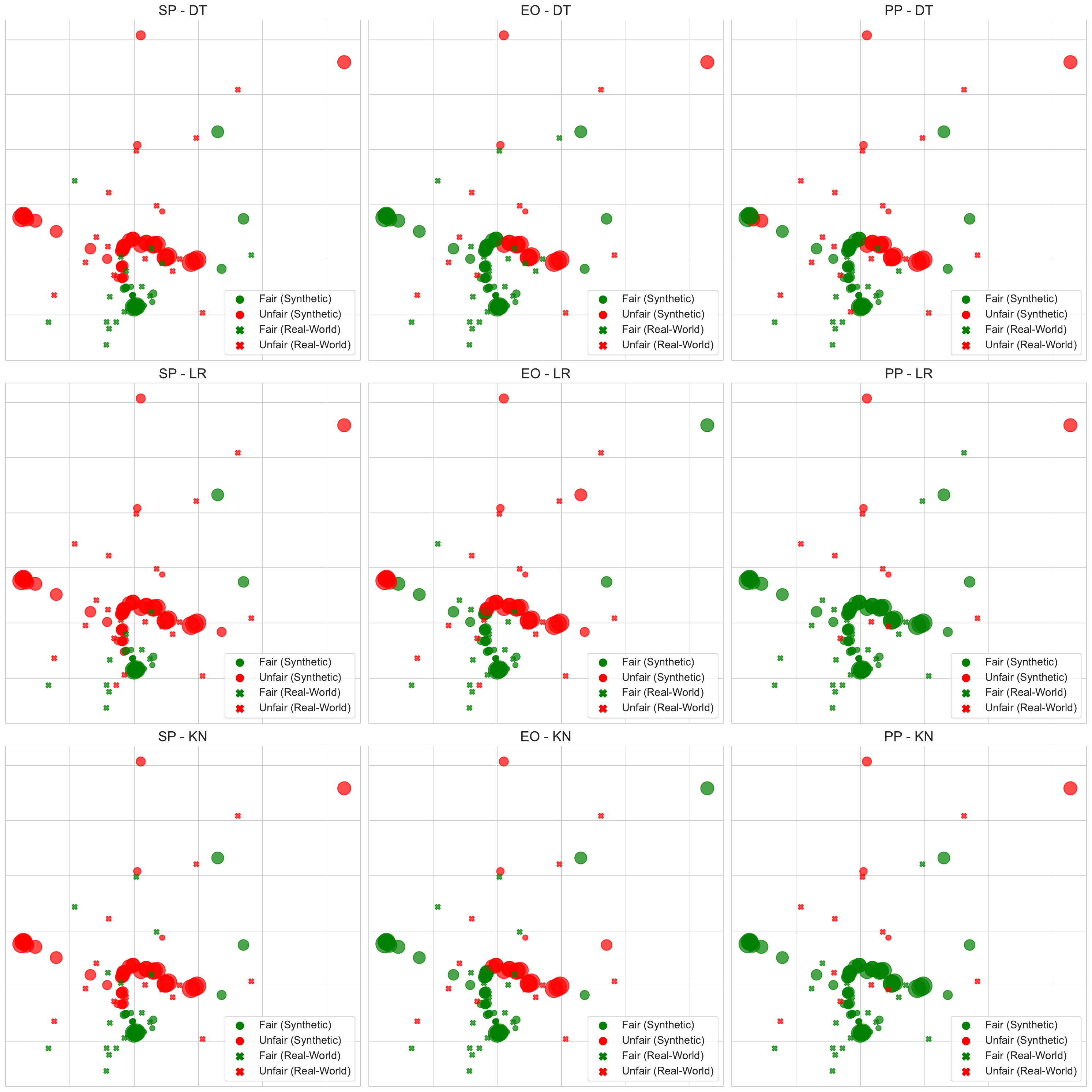}
    \caption{MDS visualization of synthetic and real-world datasets. Synthetic datasets are represented by circles while real world datasets are represented by crosses. Each subplot corresponds to a specific fairness metric: Statistical Parity (SP), Equal Opportunity (EO), and Predictive Parity (PP); evaluated for Decision Trees (DT), Logistic Regression (LR), and K-Nearest Neighbors (KN). The axis represent the principal components capturing the relative positioning of datasets based on complexity metric differences. Fair results are represented in green, while unfair results are represented in red. Synthetic and real-word data show homogeneous fairness results for datasets with similar characteristics in complexity metrics.}
    \label{fig:MDS-all}
\end{figure}

\paragraph{Combination of synthetic and real-world problems}
We mix synthetic and real-world data in Figure \ref{fig:MDS-all}. The figure presents a Multidimensional Scaling (MDS) visualization of complexity metric differences across synthetic and real-world datasets for various classification models. 

There is a separation between fair and unfair datasets, with synthetic and real-world datasets displaying different clusters, indicating structural differences in fairness distributions. The relative positions of the fairness metrics vary by classification model, suggesting that different classifiers exhibit divergent fairness behavior when exposed to real and synthetic data. Some synthetic fairness distributions align closely with real-world fairness patterns, particularly for SP and EO in Logistic Regression (LR) and Decision Trees (DT), reinforcing the relevance of synthetic dataset-based analysis. This visualization provides insights into how fairness disparities manifest across classifiers and dataset types, supporting a deeper understanding of fairness behavior in machine learning models.

\section{Discussion}
\label{lab:discussion}

Our results reveal that group-specific classification complexity differences early may be indicators of unfairness outcomes across different machine learning methods. Through the analysis of 73 synthetic datasets, each generated to represent distinct bias scenarios, and additional real-world datasets, we observed non-uniform patterns in how complexity differences arise between privileged and unprivileged groups. In particular, scenarios that involved undersampling (S1D, S1F) and historical bias (S2A, S3A, S4A) led to markedly higher differences in complexity metrics. These differences, in turn, consistently led to more substantial disparities in Statistical Parity (SP), Equal Opportunity (EO), and Predictive Parity (PP). In the subsequent sections, we explore and summarize the implications of our research findings, limitations and future directions.

\subsection{How do distinct bias-generating mechanisms affect the intrinsic complexity of classification tasks for privileged versus unprivileged groups?}

A clear outcome of our investigation is that not all bias raise complexity differences equally. Our experiments show that different types of bias (measurement, representational, historical, or omitted variable) lead to uneven increases in classification complexity differences for privileged vs. unprivileged subgroups. Specifically:

\begin{itemize}
    \item Measurement bias in R (S1B) and Omitted Variable bias (S1C R Omitted) did not inflate complexity differences.
    \item Measurement bias on Y (S1E) and Historical bias on R (S2A) or Y (S3A) show similar behavior in complexity differences.
    \item Historical Bias (S2A, S3A, S4A) introduced moderate to high subgroup complexity gaps. We often observed a surge in imbalance (C2). This pattern suggests that historically skewed outcomes affect class proportions. This pattern is also observable in S1E.
    \item Representation bias (S1D and S1F) produced the most pronounced differences in complexity.    
\end{itemize}

The existence of variations in classification complexity metrics may suggest the presence of specific types of bias. However, the absence of such variations does not necessarily imply the absence of bias within the data, as demonstrated by Scenarios S1B and S1C.

\subsection{What measurable connections exist between subgroup complexity differences and fairness metrics?}

A main objective was to investigate which complexity metrics differences between subgroups correlate most strongly with unfair outcomes (in terms of Statistical Parity, Equal Opportunity, and Predictive Parity). Notable patterns emerged:

\begin{itemize}
    \item Class Imbalance (C2) or the average density of the network (density) and Statistical Parity (SP). Association rule analysis revealed that C2 and density differences were often associated with unfairness measured as SP for all methods used.

    \item The neighborhood metric indicating the fraction of borderline points differences (N1) and Equal Opportunity (EO) or Statistical Parity (SP). Specifically for Logistic Regression (LR) and K-Nearest Neighbors (KN) demostrating how the overlap affect the results of linear and neighborhood techniques.

    \item The non-linearity of a linear classifier differences (N3) and SP and EO for the linear method (LR)

\end{itemize}

The insights suggest that the differences in complexity metrics at the subgroup level provide measurable indicators of concerns related to fairness. They also underscore that no single complexity factor exclusively drives unfairness; rather, combinations (e.g., overlap plus imbalance) often pose the greatest risks.

\subsection{How complexity differences serve as indicators for algorithmic bias, helping practitioners preemptively adjust data or model choices?}

Our findings strongly suggest that classification complexity metrics differences between privileged and unprivileged groups can serve as early warning signals for potential unfairness.

If subgroup imbalance is excessive, practitioners can explore data preprocessing techniques such as targeted oversampling, undersampling of the majority subgroup, or specialized synthetic data generation to rectify class proportions. When subgroup overlap is high, feature engineering or dimensionality reduction might help to better separate the classes.

Different classifiers respond differently to complexity, thus model selection is also critical in addressing fairness. A dataset that presents linearity separability differences between groups could drive large disparities in models like Logistic Regression. Tree-based or kernelized methods might be more equitable in such settings.

Measurement of differences in classification complexity between groups can be integrated into an automated pipeline that flags “high-risk” scenarios. This allows for immediate scrutiny of the data and the model selection. However, complexity alone does not diagnose why certain subgroups exhibit these patterns; it merely indicates that classification for those subgroups is more difficult. Therefore, direct domain knowledge and additional fairness interventions remain critical to understanding and correcting the unique bias profile of each dataset.

\subsection{How real-world data reinforced the obtained results?}

Unlike synthetic cases where each dataset primarily displayed one bias source, real data often involved multiple overlapping biases that cannot always be identified. This could obscure the direct correspondence between a single complexity metric and a fairness metric. However, when aggregated across multiple real datasets, the same association rules that emerged in synthetic data were frequently confirmed.

Real-world datasets reinforced our conclusion that imbalance and overlap indicators are predictors of potential fairness issues, although they may blend with other confounding variables. The fact that we still see meaningful connections between subgroup complexity differences and fairness in real data underscores the practical relevance of computing these metrics early in the ML pipeline.

The sensitivities of the algorithm to fairness remained consistent. Logistic Regression tended to produce more pronounced disparities in high nonlinear differences, while Decision Trees and k-Nearest Neighbors sometimes mitigated certain types of complexity-driven bias but revealed others like under class imbalance (C2).

\subsection{Limitations and Future Directions}

Despite offering strong evidence for the interplay of data complexity and fairness, some constraints remain that we list in terms of limitations and future directions:

\begin{itemize}
    \item We used a structured taxonomy of bias (e.g., historical, measurement, representation) with specific parameter settings. Real-world data often involve overlapping or more nuanced forms of bias that are not perfectly captured by any single synthetic scheme. A representation of more diversity could help to identify better patterns in complexity classification metrics differences.

    \item Our analysis addressed binary protected attributes (privileged vs. unprivileged). Future research could investigate multiattribute fairness or continuous protected attributes (e.g., age-based scenarios).

    \item Although rules help identify patterns, they do not establish strict causality. Interventions should be further validated, for example, by re-running experiments after removing or mitigating complexity sources.

    \item Some complexity metrics may become less reliable in high-dimensional spaces or large-scale data. Extending such metrics or developing new ones that are suited to modern big data scenarios is an important research path.
    
\end{itemize}

This study provides evidence that quantifying subgroup complexity can uncover early indicators of fairness challenges, helping practitioners proactively mitigate bias in classification tasks. The results justify future work on more robust complexity metrics and advanced bias generation simulations that further bridge the gap between controlled experimentation and real-world ML deployment.

\section{Conclusions}
\label{lab:conclusions}

This paper systematically explores how multiple bias parameters (historical, representational, and measurement) influence complexity differences between privileged and unprivileged groups, and subsequently affect fairness results. 

Our novel application of complexity-difference analysis demonstrates that monitoring how inherently difficult each subgroup’s classification task is can provide early warning of fairness issues. Specifically, complexity metrics that capture class imbalance (C2), boundary overlap (N1), or local density (density) emerge as robust indicators of potential unfairness. Early identification of group-specific complexity enables practitioners to select or tune models and data interventions more effectively.

We validate these findings in real-world datasets, observing that complexity-based differences mirror synthetic trends, offering practical guidance for identifying and mitigating subgroup disadvantage. From an applied perspective, this study emphasizes the importance of performing complexity audits during data preparation and early model development.

The limitations of our study include the controlled nature of synthetic dataset generation and the focus on a single protected attribute. Extensions could involve multi-attribute bias scenarios, continuous protected variables, or additional complexity metrics tailored to high-dimensional data. However, our study linking complexity differences to fairness outcomes provides a foundation upon which future studies can refine targeted interventions, helping practitioners move toward more equitable classification systems.

\newpage

\appendix

\section{Real World Datasets}

\begin{table}[ht]
    \centering
    \caption{Real-world datasets. The table show the protected attibute (PA) and its privileged value (PV) the number of rows (Rows) of each dataset, the number of attributes (Att) and lastly the reference of the source of the data used.}
    \label{tab:realdatasets}
    \begin{tabular}{lllllrrr}
        \toprule
        Dataset & PA & PV & Rows & Att & Reference \\ 
        \midrule
        academic & ge & F & 131 & 20 & \cite{student_academics_performance_467}\\ 
        adult & sex & male & 45222 & 13 &  \cite{adult_2}\\  
        arrhythmia &  sex & female & 418 & 249 & \cite{arrhythmia_5}\\ 
        catalunya & foreigner & Espanyol & 855 & 42 & \cite{cat2019}\\ 
        catalunya & sex & dona & 855 & 42 & \cite{cat2019} \\ 
        catalunya &  NatG & Spain & 855 & 42 & \cite{cat2019} \\ 
        compas & race & caucasian & 5278 & 8 & \cite{COMPAS} \\ 
        compas & sex & female & 5278 & 8 & \cite{COMPAS} \\ 
        credit & sex & male & 29623 & 14 & \cite{datam2023} \\ 
        crime & race & white & 1993 & 97 & \cite{REDMOND2002660} \\ 
        default & sex & male & 30000 & 23  & \cite{default_of_credit_card_clients_350}\\ 
        diabetes & sex & male & 46176 & 15 & \cite{datam2023} \\ 
        diabetes & race & caucasian & 46176 & 15 & \cite{datam2023} \\ 
        diabetes-w & age & $<40$ & 768 & 8 & \cite{smith1998} \\ 
        drugs & gender & female & 1885 & 12 & \cite{Fehrman2017231} \\ 
        dutch & sex & male & 60420 & 9 & \cite{datam2023} \\ 
        german & sex & male & 1000 & 9 & \cite{german_144} \\ 
        heart & sex & female & 303 & 7 & \cite{heart_disease_45}\\ 
        hrs & gender & female & 12774 & 24 & \cite{pmlr-v162-do22a} \\
        hrs & race & white & 12774 & 24 & \cite{pmlr-v162-do22a} \\ 
        lsat & race & white & 20715 & 6 & \cite{Wightman1998LSACNL} \\ 
        lsat & gender & female & 20715 & 6 & \cite{Wightman1998LSACNL} \\ 
        nursery & finance & convenient & 12960 & 8 & \cite{nursery_76} \\ 
        older-adults & sex & male & 70 & 15 & \cite{Ramnath2018} \\ 
        oulad & sex & male & 31482 & 10 & \cite{datam2023}\\ 
        parkinson & sex & male & 5875 & 18 & \cite{Tsanas2009} \\ 
        ricci & race & white & 118 & 4 & \cite{Valdivia2021} \\ 
        singles & sex & male & 2813 & 12 & \cite{Hastie2009}\\ 
        student & sex & female & 649 & 38 & \cite{Cortez20085} \\ 
        wine &  color & white & 6492 & 12 & \cite{wine_quality_186} \\ 
        \bottomrule
    \end{tabular}
\end{table}

\bibliographystyle{abbrvnat}
\setcitestyle{authoryear,open={(},close={)}} 
\bibliography{main}

\end{document}